\documentclass[letterpaper, 10 pt, conference]{ieeeconf}  

\IEEEoverridecommandlockouts                              

\usepackage{graphics} 
\usepackage{subcaption}
\usepackage{epsfig} 
\usepackage{mathptmx} 
\usepackage{times} 
\usepackage{amsmath} 
\usepackage{amssymb}  
\usepackage{booktabs}
\usepackage{multirow}
\usepackage{xcolor}

\title{\LARGE \bf
\textsc{SkillComposer}: Learning Reusable Skills for Natural-Language Robot Programming
}

\author{John Woods$^{1}$ and Hasti Seifi$^{1}$
\thanks{$^{1}$School of Computing and Augmented Intelligence,
        Arizona State University, Tempe, AZ
        {\tt\small \{jmwood23, hasti.seifi\}@asu.edu}}%
}

\begin{document}

\clubpenalty10000
\widowpenalty10000

\maketitle
\thispagestyle{empty}
\pagestyle{empty}

\begin{abstract}
Natural-language interfaces can lower the barrier to programming robots, but existing systems struggle when users request complex tasks. 
While large language models (LLMs) perform well with simple commands, they often struggle to generate code for multi-step tasks, decompose high-level instructions, or reuse prior solutions. 
We present SkillComposer, an interactive natural-language robot programming system for simulation environments that continually learns reusable program abstractions. 
SkillComposer uses a generate-test architecture in which an LLM iteratively generates and revises robot programs before execution. 
Successful programs are stored and processed by an online library-learning algorithm that compresses recurring function sequences into reusable macro skills for future tasks. 
We evaluate SkillComposer through ablation experiments and a user study with 12 participants to determine its effectiveness on manipulation and robot caregiving tasks. 
The results show that evaluator-guided generation and learned abstractions improve success rates and usability while reducing user effort in natural-language robot programming.
\end{abstract}

\section{Introduction}

Robotics simulators increase the safety and efficiency of designing, testing, and refining robot behaviors before deployment in the physical world \cite{todorov2012mujoco, NVIDIA_Isaac_Sim, coppeliaSim}. 
However, simulators still require users to understand programming and robotics, creating a barrier for domain experts who know what behavior they want but lack the technical expertise to implement it.
This challenge is especially visible in assistive robotics, where caregivers, clinicians, and researchers may want to specify tasks involving object manipulation, feeding, or other human-robot interaction (HRI) tasks \cite{nanavati2023physically}. 
RCareWorld \cite{ye_rcare_2022}, a Unity-based caregiving robot simulator with Python and Robot Operating System (ROS) interfaces, provides a useful testbed for these tasks, but specifying robot behavior remains a programming-heavy process.

Recent advances in Large Language Models (LLMs) have made natural-language robot programming increasingly practical, allowing users to describe tasks in ordinary language rather than writing low-level robot code \cite{vemprala_chatgpt_2024, singh_progprompt_2023, ahn_as_2022}. 
Prior systems have shown that LLMs can select robot actions, generate executable programs, and help non-expert users specify robotic tasks \cite{karli_alchemist_2024, ge_cocobo_2024}. 
However, these systems typically rely on a fixed set of manually defined primitive skills, which can require users to decompose complex goals, repair failed generations, or adapt their language to the system's available API. 
Most also treat each prompt as an independent interaction, rather than learning reusable skills from recurring program structure.

To address these limitations, we introduce SkillComposer, an interactive natural-language robot programming system that learns reusable robot skills from generated programs. 
Rather than treating each user prompt as an isolated generation task, the system generates, evaluates, and stores successful robot programs for future reuse.
A generate-test loop uses a coder LLM to produce candidate programs and an evaluator LLM to provide feedback before a program is accepted. 
Accepted programs are added to an experience buffer and processed by the online library-learning algorithm Stitch \cite{bowers_top-down_2023}, which identifies recurring code structures and compresses them into reusable macro skills. 
These macros are provided to the coder model for subsequent prompts, allowing the system to reuse learned skills instead of repeatedly generating the same low-level function sequences.

To assess SkillComposer, we ask two research questions: 
\textbf{RQ1:} Does combining generate-test with library learning improve performance across manipulation and interaction tasks? 
\textbf{RQ2:} How does SkillComposer affect usability and user experience compared to a baseline coding LLM?
We answer RQ1 with ablation experiments across four system variants in generic manipulation and assisted feeding scenes, measuring task success, program length, runtime, and macro usage.
To answer RQ2, we conduct a user study with 12 participants comparing a baseline coding LLM against SkillComposer.
We measure task completion rate and time as participants complete object rearrangement, meal preparation, and care-recipient feeding tasks, followed by usability ratings and qualitative feedback.

Overall, this paper contributes: 
(i) SkillComposer, an interactive natural-language robot programming system that learns reusable robot skills from generated robot programs, and
(ii) empirical results from ablation experiments and a user study demonstrating how generate-test and online skill learning affect task success, efficiency, skill reuse, and usability in generic manipulation and assisted feeding scenarios.
Together, these contributions show how library-learning techniques from programming languages research can enable natural language robot programming systems to move beyond one-off code generation toward interfaces that retain, reuse, and refine task knowledge over repeated use.

\section{Related Work}

\subsection{Natural Language Robot Programming}

LLMs have increasingly been explored as natural language interfaces for robot programming.
SayCan selects robot actions from language instructions based on the robot's physical capabilities \cite{ahn_as_2022}, while ProgPrompt generates robot programs from available actions, objects, and example programs \cite{singh_progprompt_2023}.
ChatGPT for Robotics demonstrates how general-purpose LLMs can generate robot code and task plans when provided with the robot API in the prompt \cite{vemprala_chatgpt_2024}, and GenSim uses LLMs to generate code for robotic tasks and saves high-quality generations to a library for future model fine-tuning and evaluation \cite{wang2024gensim}.
Recent systems have also focused on making robot programming more accessible for end users.
Alchemist uses LLMs to enable novice and expert users to program robot pick-and-place tasks using natural language \cite{karli_alchemist_2024}, but they report that LLM-generated code is often unreliable and that effective prompting remains challenging for novices. 
Cocobo pairs LLMs with an interactive diagram interface to enable modification and debugging of programs \cite{ge_cocobo_2024}, although some users found the interface challenging to use without programming knowledge.
Together, these systems demonstrate the potential of natural language as an interface for robot programming, but they generally rely upon a fixed set of manually provided skills \cite{vemprala_chatgpt_2024, singh_progprompt_2023, ahn_as_2022}. 
In contrast, SkillComposer focuses on enabling the programming interface to improve over time by learning reusable macro skills from previously generated programs.

Several works have also investigated methods for improving the reliability of LLM-generated code. 
Code-generation benchmarks emphasize that generated programs should be evaluated for task correctness in addition to syntactic validity \cite{10685214}. 
MultiTalk uses a generate-test loop in which a planning LLM iteratively refines generated robot programs based on feedback from an analyzer LLM and visual input \cite{11128486}.
SkillComposer builds on this line of work through a generate-test loop that combines LLM-based functional evaluation with deterministic syntax validation before execution. 
Unlike prior approaches \cite{11128486}, accepted programs are retained and used as a growing corpus for online macro learning, allowing the system to improve future generations rather than only the current one.

\subsection{Learning Reusable Robot Skills and Program Abstractions}

Many recent studies have explored how agents and robots can expand their capabilities through reusable tools or skills.
In LLM agent settings, LATM \cite{cai_large_2024} and CREATOR \cite{qian_creator_2023} use LLMs to generate new callable tools via prompting rather than extracting them from previously generated programs, while Voyager continually builds a library of executable skills through interaction with the Minecraft environment \cite{wang_voyager_2023}.
In robotics, LLRL uses LLMs and a wake-sleep paradigm to build a library of reusable manipulation skills with guidance from a human teacher \cite{tziafas_lifelong_2024}, and HALP uses an LLM to identify when robots should learn new skills from human interaction \cite{parakh_lifelong_2024}.
LYRA extends LLM-based robot task generation to long-horizon settings, but relies on users to design a learning curriculum and guide skill acquisition \cite{meng_growing_2025}.
While prior robotics systems demonstrate that reusable skills can support long-horizon task execution, skill acquisition often depends on explicit human teaching, curriculum design, or interaction with the environment \cite{tziafas_lifelong_2024, parakh_lifelong_2024, meng_growing_2025}.
In contrast, SkillComposer learns reusable robot skills directly from accepted LLM-generated robot programs, allowing the callable skill set to grow from the user's interaction history without requiring users to manually design new skills.

Outside of robotics, research in the programming languages community investigates how systems can learn reusable abstractions from existing programs. 
DreamCoder learns reusable library functions from a corpus of programs using a wake-sleep paradigm to improve future program synthesis \cite{ellis_dreamcoder_2023}.
Stitch later introduced an efficient library learning algorithm that discovers abstractions to minimize program corpus size \cite{bowers_top-down_2023}.
Others have combined library learning with language models.
LAPS builds upon DreamCoder and uses natural language to guide abstraction learning and program synthesis \cite{wong_leveraging_2021}, while LILO combines Stitch with LLM-guided synthesis to create documentation of learned abstractions \cite{grand_lilo_2024}.
Yet, these approaches focus on functional program synthesis over fixed corpora, whereas SkillComposer adapts library learning to imperative robot programs generated during interactive natural-language programming.

\begin{figure*}[htbp]
\centering
\includegraphics[width=0.9\linewidth]{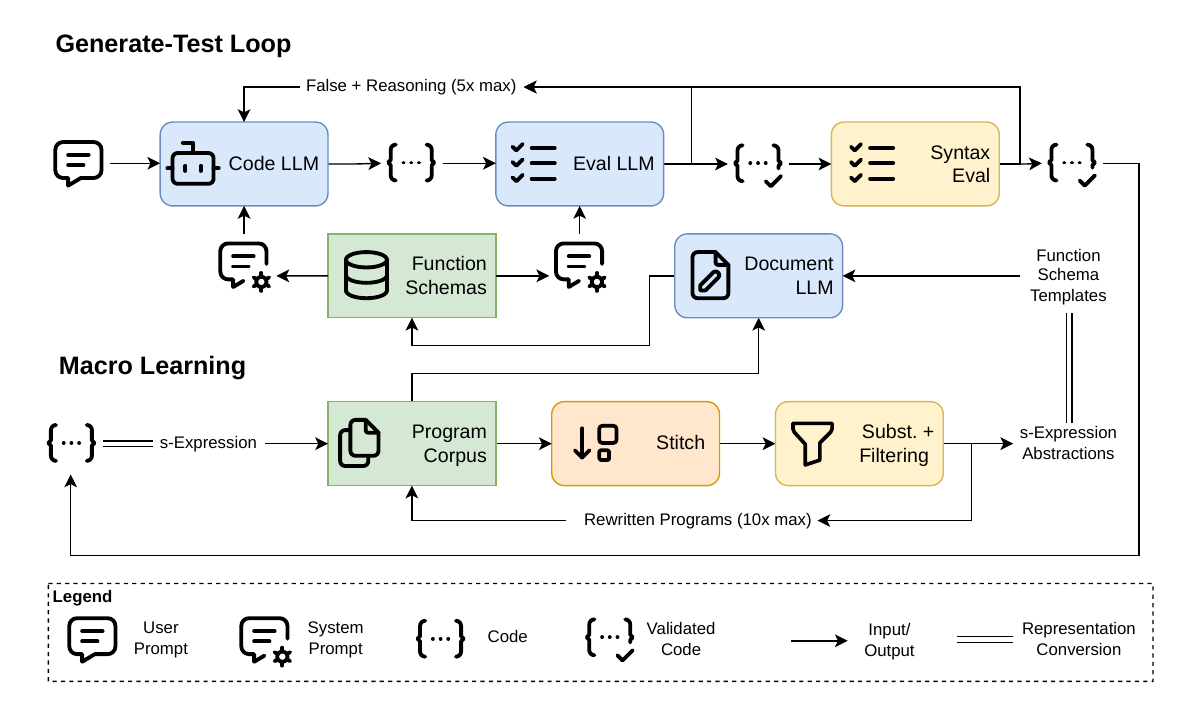}
\caption{An overview of the SkillComposer architecture. Blue modules are natural language models, the orange module is the library learning algorithm, yellow modules are static algorithms, and green modules are data structures.}
\label{fig:architecture}
\end{figure*}

\begin{figure}[htbp]
\centering
\includegraphics[width=1.0\linewidth]{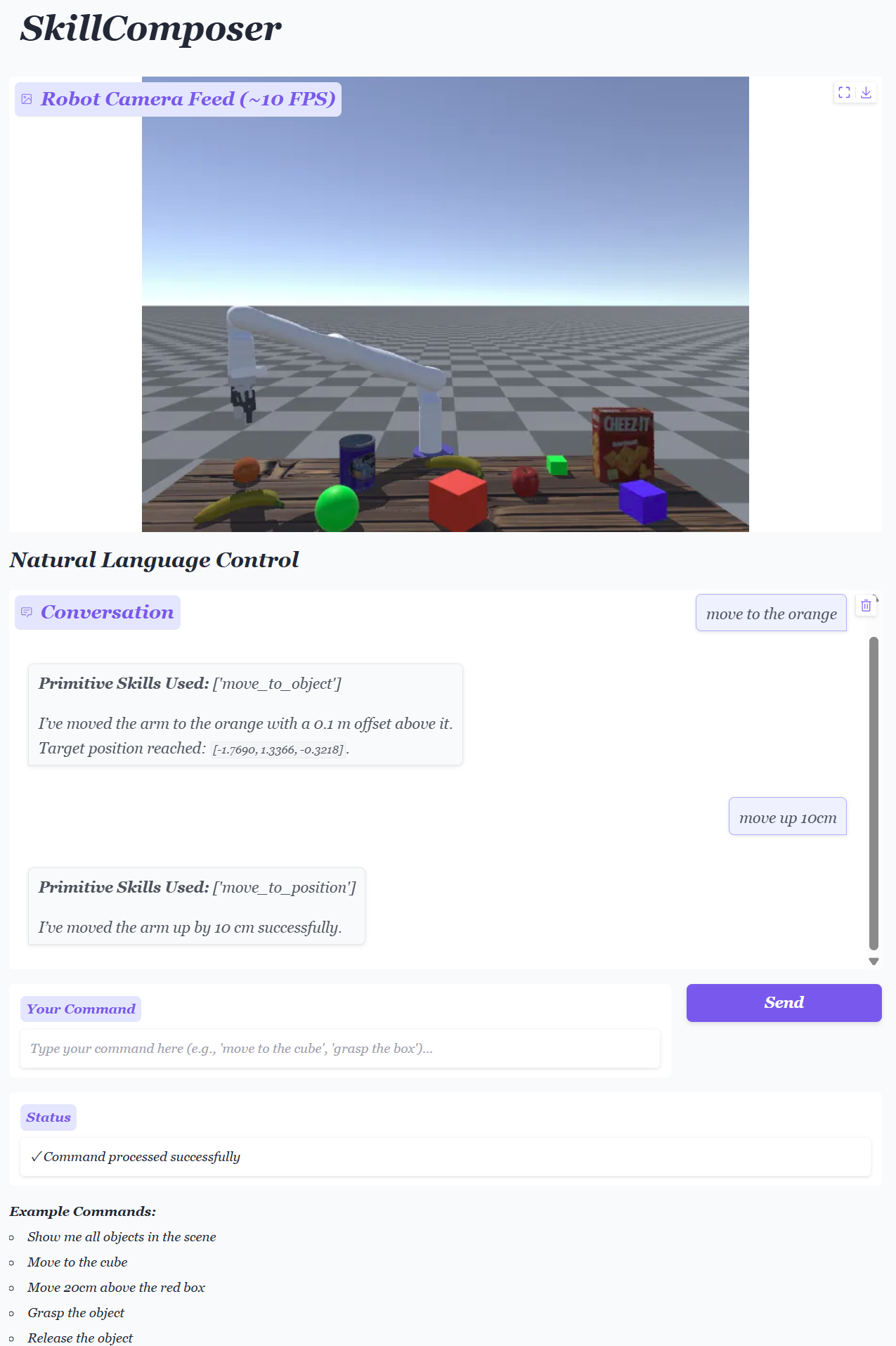}
\caption{The SkillComposer user interface, which contains a live camera feed from the simulation and a chat interface for interacting with the system.}
\label{fig:ui}
\end{figure}

\begin{figure}[htbp]
\centering
\includegraphics[width=1.0\linewidth]{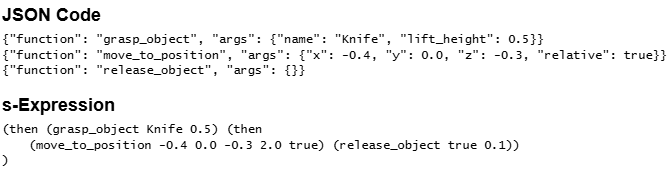}
\caption{An equivalent portion of code in the JSON representation (top) and s-expression representation (bottom). Default arguments are added into the s-expression based on the defaults listed in the function schema.}
\label{fig:code}
\end{figure}

\section{SkillComposer}

SkillComposer is an interactive natural language robot programming system that combines two mechanisms:
(i) a generate-test loop for producing executable robot programs from user prompts, and
(ii) an online macro learning module that discovers reusable skills from previously accepted programs.
The full architecture is shown in Figure \ref{fig:architecture}.
Users interact with the system through a locally hosted Gradio interface, shown in Figure \ref{fig:ui}, which supports natural language prompting and manual robot and camera control.
The interface displays the RCareWorld \cite{ye_rcare_2022} simulation and conversation history, which includes a summary of generated function calls and a natural-language explanation after each prompt, allowing users to inspect which skills the system uses.

\subsection{Generate-Test Loop} 
The Generate-Test Loop includes a coder LLM to generate candidate robot programs and a two-stage evaluation pipeline for checking functional and structural validity of programs before execution.

\paragraph{Coder LLM} 
Given a user instruction, SkillComposer first sends the prompt to the coder LLM, which generates a candidate robot program from the available function schemas.
The system starts out with five primitive functions: \verb|get_info|, \verb|move_to_object|, \verb|move_to_position|, \verb|grasp_object|, and \verb|release_object|.
Programs are represented as JSON function calls, where each call specifies a function name and a dictionary of argument values.
This structured representation allows generated programs to be parsed, validated, and executed as robot actions.
Using schema-constrained JSON distinguishes SkillComposer from approaches that ask LLMs to generate free-form code, since generated programs can be easily checked against the available robot API before execution.

\paragraph{Evaluator Module} 
The generated program is evaluated in two stages.
First, the candidate program and user prompt are provided to an evaluator LLM.
The evaluator judges whether the program satisfies the user's request and, if not, returns natural language feedback describing the suspected problem.
Programs that pass this stage are then checked by a static validator, which verifies that the program is well-formed, uses functions from the available schema set, provides valid arguments, and includes all required parameters.
This ensures the final LLM output is structurally correct and enables the evaluator LLM to focus solely on program functionality.

\paragraph{Program Validation and Refinement} 
A program is accepted if it passes both the evaluator LLM and static validator.
Accepted programs are executed in RCareWorld before being sent on to the macro learning module.
If either evaluation stage fails, the feedback is appended to the coder LLM conversation and the coder is prompted to revise the program.
This process repeats for up to five iterations, after which the system reports an error to the user rather than executing an invalid program.

\subsection{Online Macro Learning}
SkillComposer learns reusable robot skills by adapting the library learning algorithm Stitch \cite{bowers_top-down_2023} for online learning with the imperative robot programs generated during interaction.

\paragraph{Generating Macro Candidates with Stitch}
To apply Stitch to robot programs, SkillComposer converts each accepted JSON program from the generate-test loop into a symbolic expression (s-expression).
Each primitive action becomes a positional function call, with arguments ordered by the corresponding function schema.
Sequential execution is represented with a right-nested \verb|then| structure, preserving the imperative order of robot actions while producing a tree-structured format suitable for library learning.
An example conversion is shown in Figure \ref{fig:code}.
Each accepted s-expression is appended to an initially empty corpus of previously generated robot programs.
Unlike standard library-learning algorithms that operate over a fixed corpus, this corpus grows as the user interacts with the system.

SkillComposer applies Stitch to this corpus to learn reusable robot skills.
Stitch iteratively searches for abstractions that reduce the total size of a program corpus when the corpus is rewritten to use those abstractions.
In SkillComposer, this objective is used to identify repeated sequences of robot function calls that can be compressed into higher-level macro skills.
We run Stitch for a maximum of 20 iterations and with a maximum arity of five to generate a diverse set of candidate abstractions.

\paragraph{Substitution and Filtering} 
Because Stitch was designed for functional languages rather than imperative robot programs, its raw abstractions often cannot be directly converted to robot skills.
The key adaptation is to enforce that learned abstractions are valid robot action sequences, rather than arbitrary expressions that merely compress the corpus.
Due to Stitch's iterative nature, later abstractions may reference abstractions learned earlier in the process.
Abstraction calls are therefore recursively inlined so that each candidate macro is expressed only in terms of primitive robot functions.
This substitution step helps maximize the amount of valid candidate abstractions before filtering. 
The expansion procedure also introduces fresh variables for any missing arguments, converting partial function applications learned by Stitch into valid robot programs.

The resulting abstractions are then checked for structural validity.
A valid macro must contain at least two primitive function calls, use \verb|then| only as a binary sequencing operator, use only known primitive functions, and place variables only in primitive argument positions.
These constraints filter out abstractions that may be syntactically valid for Stitch but do not correspond to executable imperative robot programs.
Finally, variables are renamed consistently so duplicate abstractions can be identified and removed.

Because many high-ranking Stitch abstractions are invalid under these constraints, SkillComposer reimplements the outer Stitch learning loop to select only valid abstractions at each iteration.
After filtering, SkillComposer selects the highest-ranked valid candidate and rewrites the corpus using that abstraction.
This repeats until either ten valid abstractions have been learned or Stitch produces no additional valid candidates.
The macro set is relearned after each accepted prompt, so the available skills always reflect the user's full interaction history.

\paragraph{Generating a Function Schema} 
Each valid abstraction is converted into a function schema for use with the coder and evaluator LLMs.
The schema specifies the macro name, description, parameters, parameter types, and parameter descriptions.
Parameter names and types are inferred by tracing abstraction variables to their corresponding primitive function arguments.
The schema template is then sent to a documentation LLM, along with the macro body and contextual examples from user prompts and generated programs in which the abstraction appears.
The documentation LLM generates a human-readable skill name, skill description, and parameter descriptions, which are added to the macro schema.
This allows the schema to reflect both the learned program structure and the user's natural-language prompts.

\paragraph{Using Learned Macros} 
The completed schema is added to the same function schema store as primitive functions, so future coder and evaluator prompts include both primitives and learned macros.
When generating future programs, the coder LLM may call a learned macro if deemed relevant to the user's prompt.
Before execution, macro calls are expanded back into primitive robot calls, preserving simulator compatibility while allowing the LLM to generate programs at a higher level of abstraction.
After execution, the conversation UI shows the skills used and a natural language summary of the output.

\begin{figure*}[ht]
    \centering
    \begin{subfigure}[b]{0.32\linewidth}
        \includegraphics[width=\linewidth]{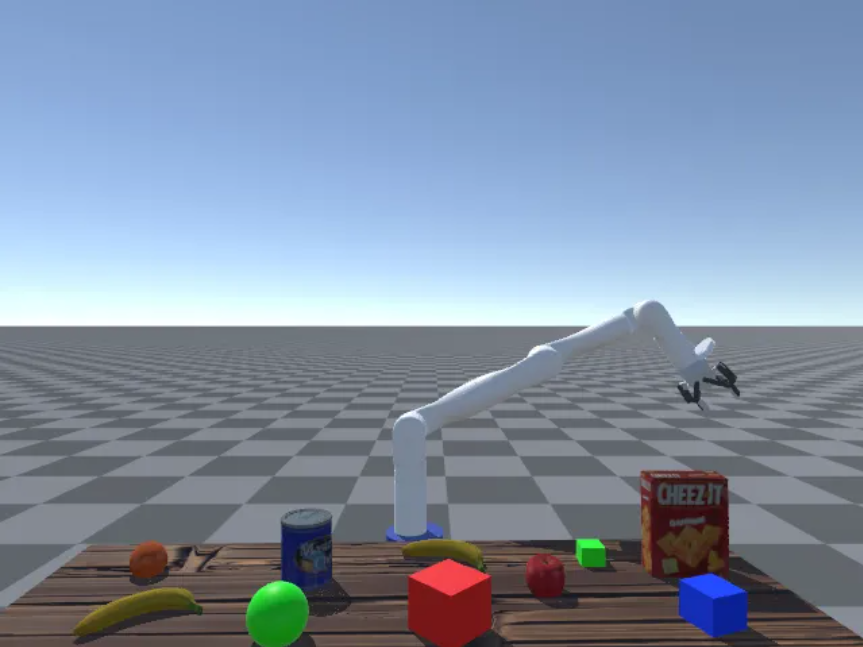}
        \caption{}
        \label{fig:env_objects}
    \end{subfigure}
    \begin{subfigure}[b]{0.32\linewidth}
        \includegraphics[width=\linewidth]{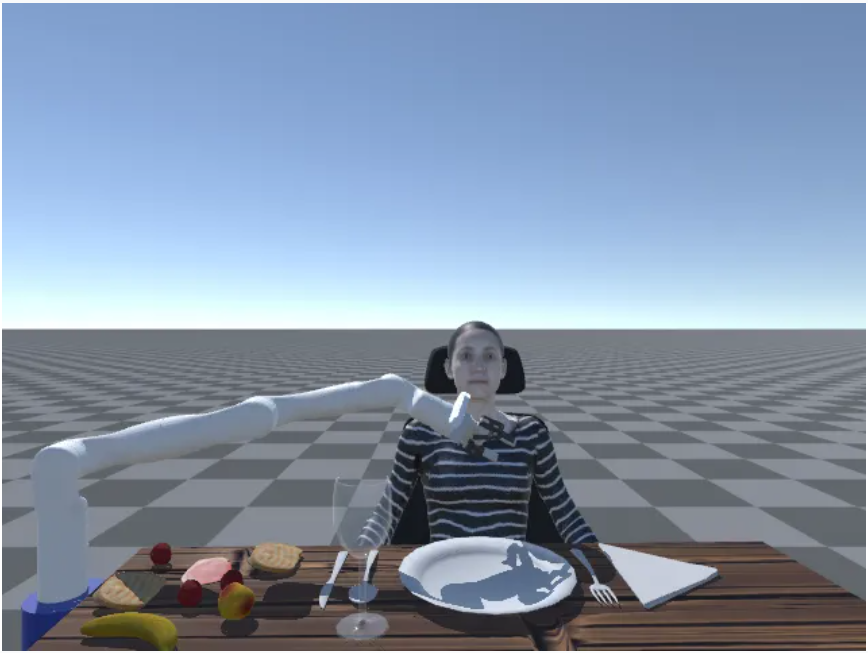}
        \caption{}
        \label{fig:env_feeding}
    \end{subfigure}
    \begin{subfigure}[b]{0.32\linewidth}
        \includegraphics[width=\linewidth]{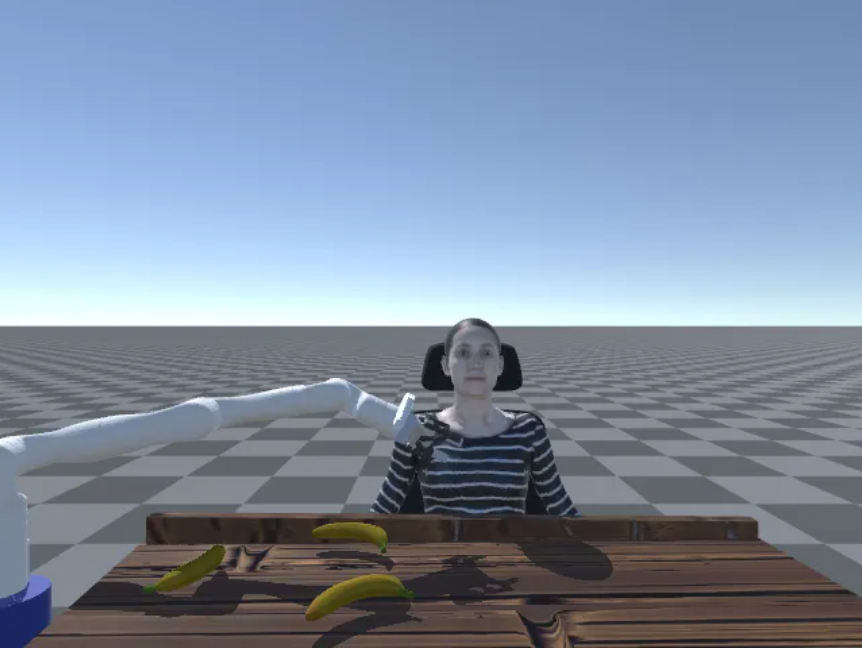}
        \caption{}
        \label{fig:env_bananas}
    \end{subfigure}
    \caption{Three simulation environments used in the experiments. (a) Objects: evaluates object manipulation, spatial reasoning, multi-step actions, and goal planning (prompts 1–20). (b) Feeding: evaluates simple and multi-step HRI tasks (prompts 21–40) and is used in the main user study. (c) Bananas: used for the user study practice task.}
    \label{fig:envs}
\end{figure*}

\section{Experiments}

We evaluate SkillComposer through two complementary studies. 
First, we conduct ablation experiments to isolate the effects of the generate-test loop and online macro learning on task performance (RQ1). 
Second, we conduct a within-subjects study to compare the user experience of SkillComposer against a baseline coding LLM in an interactive robot programming setting (RQ2).

\subsection{Ablation Experiments}

To evaluate the contribution of each component of SkillComposer (RQ1), we perform ablation experiments across four system variants: (1) \emph{Coder LLM} where the LLM generates robot programs directly from user prompts, 
(2) \emph{Coder LLM + Eval LLM} includes the generate-test loop, allowing candidate programs to be revised using evaluator feedback before execution, 
(3) \emph{Coder LLM + Stitch} includes online macro learning, allowing the coder LLM to use previously learned macro skills, but does not use evaluator-guided revision, and 
(4) \emph{SkillComposer} combines the generate-test loop with online macro learning.

Each variant is evaluated in two RCareWorld environments. 
The Objects environment evaluates basic object manipulation, spatial reasoning, multi-step action planning, and goal planning (Figure \ref{fig:env_objects}). 
The Feeding environment evaluates assistive caregiving tasks involving both simple and multi-step Human-Robot Interaction (Figure \ref{fig:env_feeding}). 
For each environment, we use a set of 20 natural-language prompts covering the aforementioned areas.

We measure four dependent variables: (1) \emph{success rate} measures whether the generated program successfully completes the requested prompt in simulation, (2) \emph{program length} counts the number of function calls in the final generated program as a measure of program complexity, (3) \emph{macro usage} measures the percentage of program length that consists of macro calls, and (4) \emph{runtime} measures the time required to produce an accepted program, including any generate-test iterations when applicable. 
Because LLM outputs are stochastic, each system-environment pair is run five times, and results are averaged across trials.

\begin{table*}[t]
\centering
\small
\caption{Summary of ablation experiments for each prompt category, averaged over all trials and environments. A representative prompt from each category is provided for reference.}
\label{tab:ablation_results}
\resizebox{\textwidth}{!}{%
\begin{tabular}{lllcccc}
\toprule
Prompt Category & Example Prompt & System & Success Rate & Program Length & Macro Usage (\%) & Runtime (s) \\
\midrule
\multirow{4}{*}{Object Manipulation} & \multirow{4}{*}{\parbox{3cm}{\raggedright\emph{Grab a yellow fruit}}} & Coder LLM & \textbf{1.00 $\pm$ 0.00} & 1.00 $\pm$ 0.00 & -- & 2.26 $\pm$ 0.80 \\
 &  & Coder LLM + Eval LLM & 0.96 $\pm$ 0.09 & 1.32 $\pm$ 0.44 & -- & 9.58 $\pm$ 9.86 \\
 &  & Coder LLM + Stitch & \textbf{1.00 $\pm$ 0.00} & 1.08 $\pm$ 0.11 & 0.00 $\pm$ 0.00 & \textbf{2.24 $\pm$ 0.67} \\
 &  & SkillComposer & 0.96 $\pm$ 0.09 & 1.32 $\pm$ 0.36 & \textbf{8.00 $\pm$ 17.89} & 13.97 $\pm$ 9.80 \\
\midrule
\multirow{4}{*}{Spatial Reasoning} & \multirow{4}{*}{\parbox{3cm}{\raggedright\emph{Move rightmost banana to the left of the can}}} & Coder LLM & 0.24 $\pm$ 0.43 & 0.96 $\pm$ 0.41 & -- & \textbf{11.94 $\pm$ 4.05} \\
 &  & Coder LLM + Eval LLM & \textbf{0.76 $\pm$ 0.43} & 2.82 $\pm$ 0.76 & -- & 50.43 $\pm$ 17.02 \\
 &  & Coder LLM + Stitch & 0.24 $\pm$ 0.43 & 1.00 $\pm$ 0.58 & 0.00 $\pm$ 0.00 & 14.51 $\pm$ 5.80 \\
 &  & SkillComposer & 0.40 $\pm$ 0.42 & 2.06 $\pm$ 0.90 & \textbf{37.33 $\pm$ 43.87} & 30.47 $\pm$ 18.28 \\
\midrule
\multirow{4}{*}{Multi-Step Actions} & \multirow{4}{*}{\parbox{3cm}{\raggedright\emph{Pick up the sphere, move it to the right 50cm, then release it}}} & Coder LLM & 0.32 $\pm$ 0.27 & 1.68 $\pm$ 0.46 & -- & \textbf{7.21 $\pm$ 2.55} \\
 &  & Coder LLM + Eval LLM & \textbf{1.00 $\pm$ 0.00} & 3.60 $\pm$ 1.52 & -- & 22.51 $\pm$ 11.81 \\
 &  & Coder LLM + Stitch & 0.32 $\pm$ 0.23 & 1.80 $\pm$ 0.71 & 0.00 $\pm$ 0.00 & 9.38 $\pm$ 3.28 \\
 &  & SkillComposer & 0.96 $\pm$ 0.09 & 2.32 $\pm$ 1.10 & \textbf{45.33 $\pm$ 51.08} & 20.59 $\pm$ 10.74 \\
\midrule
\multirow{4}{*}{Goal Planning} & \multirow{4}{*}{\parbox{3cm}{\raggedright\emph{Stack all the cubes}}} & Coder LLM & 0.08 $\pm$ 0.18 & 1.72 $\pm$ 2.06 & -- & \textbf{11.67 $\pm$ 3.62} \\
 &  & Coder LLM + Eval LLM & 0.68 $\pm$ 0.39 & 13.07 $\pm$ 11.21 & -- & 50.33 $\pm$ 26.83 \\
 &  & Coder LLM + Stitch & 0.00 $\pm$ 0.00 & 0.64 $\pm$ 0.22 & 0.00 $\pm$ 0.00 & 16.50 $\pm$ 5.07 \\
 &  & SkillComposer & \textbf{0.76 $\pm$ 0.33} & 6.17 $\pm$ 4.48 & \textbf{86.00 $\pm$ 21.91} & 49.18 $\pm$ 9.25 \\
\midrule
\multirow{4}{*}{Human-Robot Interaction} & \multirow{4}{*}{\parbox{3cm}{\raggedright\emph{Put the fork up to the user's mouth}}} & Coder LLM & 0.14 $\pm$ 0.19 & 1.16 $\pm$ 0.32 & -- & 15.96 $\pm$ 4.33 \\
 &  & Coder LLM + Eval LLM & \textbf{0.94 $\pm$ 0.13} & 3.44 $\pm$ 1.05 & -- & 47.49 $\pm$ 15.50 \\
 &  & Coder LLM + Stitch & 0.16 $\pm$ 0.18 & 1.16 $\pm$ 0.23 & 4.00 $\pm$ 8.43 & \textbf{10.71 $\pm$ 2.98} \\
 &  & SkillComposer & 0.86 $\pm$ 0.23 & 2.86 $\pm$ 1.04 & \textbf{22.00 $\pm$ 23.94} & 33.36 $\pm$ 12.14 \\
\midrule
\multirow{4}{*}{Multi-Step Human-Robot Interaction} & \multirow{4}{*}{\parbox{3cm}{\raggedright\emph{Feed the peach first and then the banana to the user}}} & Coder LLM & 0.02 $\pm$ 0.06 & 1.46 $\pm$ 1.07 & -- & 16.06 $\pm$ 2.50 \\
 &  & Coder LLM + Eval LLM & 0.98 $\pm$ 0.06 & 10.39 $\pm$ 6.05 & -- & 53.95 $\pm$ 11.09 \\
 &  & Coder LLM + Stitch & 0.04 $\pm$ 0.08 & 1.52 $\pm$ 0.90 & 5.00 $\pm$ 10.54 & \textbf{11.77 $\pm$ 3.60} \\
 &  & SkillComposer & \textbf{1.00 $\pm$ 0.00} & 5.50 $\pm$ 1.75 & \textbf{55.82 $\pm$ 34.06} & 27.98 $\pm$ 11.37 \\
\bottomrule
\end{tabular}%
}
\end{table*}

\begin{figure*}[htbp]
\centering
\includegraphics[width=1.0\linewidth]{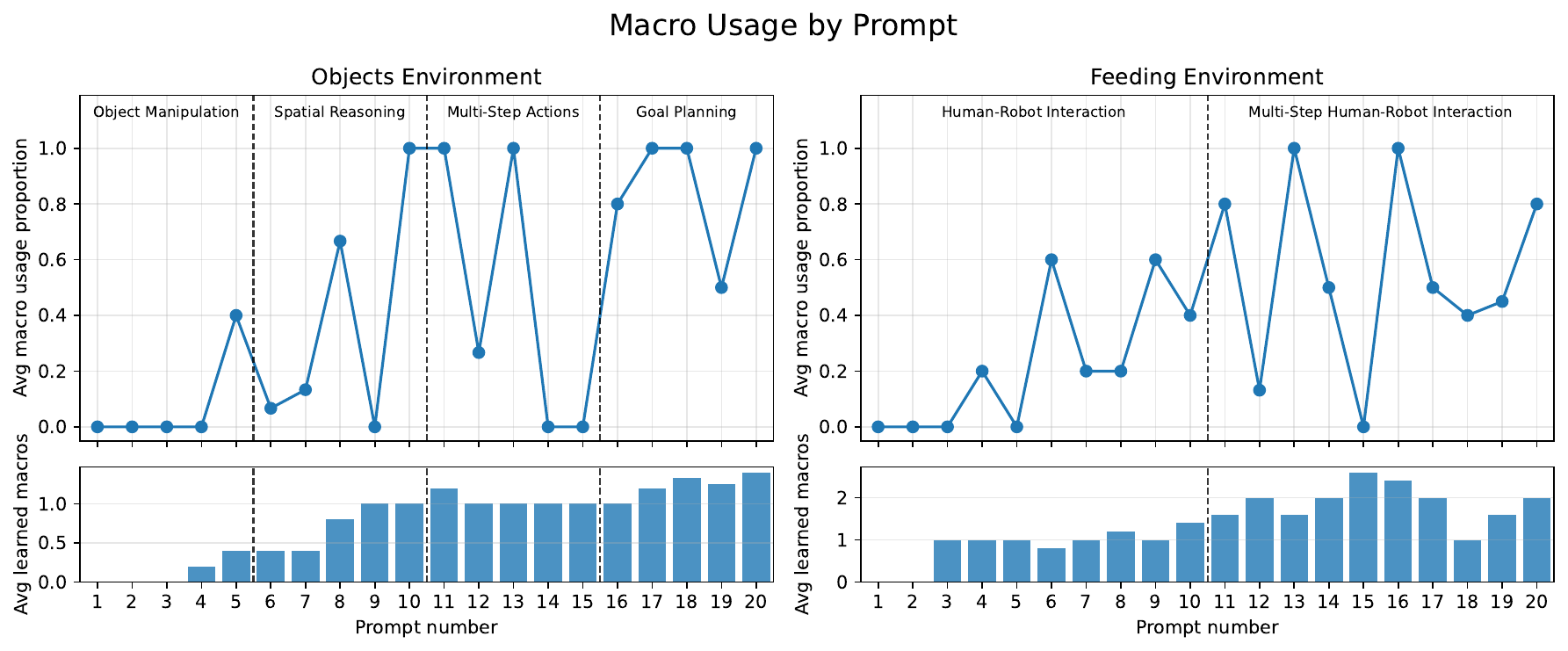}
\caption{Top: Average SkillComposer macro usage across five runs in the Objects (left) and Feeding (right) environments versus the number of input prompts. Bottom: Average number of learned macros across five runs versus input prompts.}
\label{fig:macro}
\end{figure*}

\subsection{User Study}

To evaluate the user experience of SkillComposer (RQ2), we ran a within-subjects user study comparing the full SkillComposer system with a baseline coding LLM, similar to the one used in prior work \cite{chen_rcaregen_2025}. 
The baseline condition consisted of the coder LLM, while the SkillComposer condition included both the generate-test loop and online macro learning. 
The user-facing interface was identical across conditions, and backend order was counterbalanced across participants to reduce order effects.
Participants received a \$15 Amazon gift card for their time. The study was approved by the institutional review board (IRB).

We recruited 12 participants (3 female, 9 male, 23-67 years old). Participants had upper-intermediate English proficiency, normal or corrected-to-normal vision, and none or beginner-level experience with robotics or programming. 
These criteria targeted novice end users likely to benefit from natural-language robot programming.

Each study session lasted about 75 minutes. 
Participants first received an overview of the study, provided written consent, and completed a background questionnaire collecting demographic information and prior experience with robotics and programming. 
They then completed a short practice task (moving a banana 25 cm to the right) in the Bananas environment (Figure \ref{fig:env_bananas}) to become familiar with the interface.

Next, participants completed tasks using both systems. 
For each system, participants worked in the Feeding environment and completed three high-level robot programming tasks: rearranging objects on the table, preparing a meal, and feeding the care recipient. 
Participants wrote their own natural-language prompts and could send as many prompts as desired.
We collected both objective and subjective measures. 
Objective measures included the number of prompts used, task success rate, and task completion time. 
These metrics assessed how well participants could complete the above high-level tasks using the system. 
Subjective measures included the System Usability Scale (SUS) \cite{brooke_sus_1996} and 7-point Likert ratings of the system's effectiveness in understanding commands, output predictability, alignment of robot skills with the task, and overall satisfaction.

At the end, participants engaged in a semi-structured interview. 
The interview asked participants to compare the two systems, describe which system they preferred, identify aspects that were helpful or confusing, and discuss whether either system seemed to adopt or reuse useful behaviors over time. 
Participants were also asked about the meaningfulness of learned skills, prompting strategies they used, types of tasks the systems handled well or poorly, and suggestions for improvement. 
The interview responses were audio-recorded and later transcribed for analysis.

\section{Results}

\subsection{Ablation Experiments}

Table \ref{tab:ablation_results} summarizes the ablation results for each prompt category.
The generate-test loop produced the largest improvement in success rate, with evaluator-guided systems outperforming direct code generation across most categories.
The exception was basic object manipulation, where all variants performed well, suggesting that direct LLM generation is often sufficient for simple commands.
However, the generate-test loop also introduced the largest latency cost, increasing generation time by approximately four to five times due to additional evaluation and revision steps.

\begin{table*}[ht!]
\centering
\small
\caption{Summary of user study task metrics, averaged over all participants.}
\label{tab:user_results}
\begin{tabular}{llccc}
\toprule
System & Task & Success Rate & Num. Prompts & Task Time (m:ss) \\
\midrule
\multirow{3}{*}{Baseline} & Rearrange Objects & 0.50 $\pm$ 0.52 & 5.8 $\pm$ 2.2 & 6:09 $\pm$ 2:57 \\
 & Meal Preparation & \textbf{0.92 $\pm$ 0.29} & 8.2 $\pm$ 5.6 & 6:10 $\pm$ 3:19 \\
 & Feed Care Recipient & \textbf{0.75 $\pm$ 0.45} & 6.6 $\pm$ 2.7 & \textbf{3:48 $\pm$ 1:43} \\
\midrule
\multirow{3}{*}{SkillComposer} & Rearrange Objects & \textbf{0.92 $\pm$ 0.29} & \textbf{1.9 $\pm$ 1.2} & \textbf{3:20 $\pm$ 1:20} \\
 & Meal Preparation & \textbf{0.92 $\pm$ 0.29} & \textbf{3.2 $\pm$ 2.6} & \textbf{3:59 $\pm$ 1:48} \\
 & Feed Care Recipient & 0.67 $\pm$ 0.49 & \textbf{3.5 $\pm$ 1.6} & 4:16 $\pm$ 1:57 \\
\bottomrule
\end{tabular}
\end{table*}

\begin{figure}[htbp]
\centering
\includegraphics[width=1.0\linewidth]{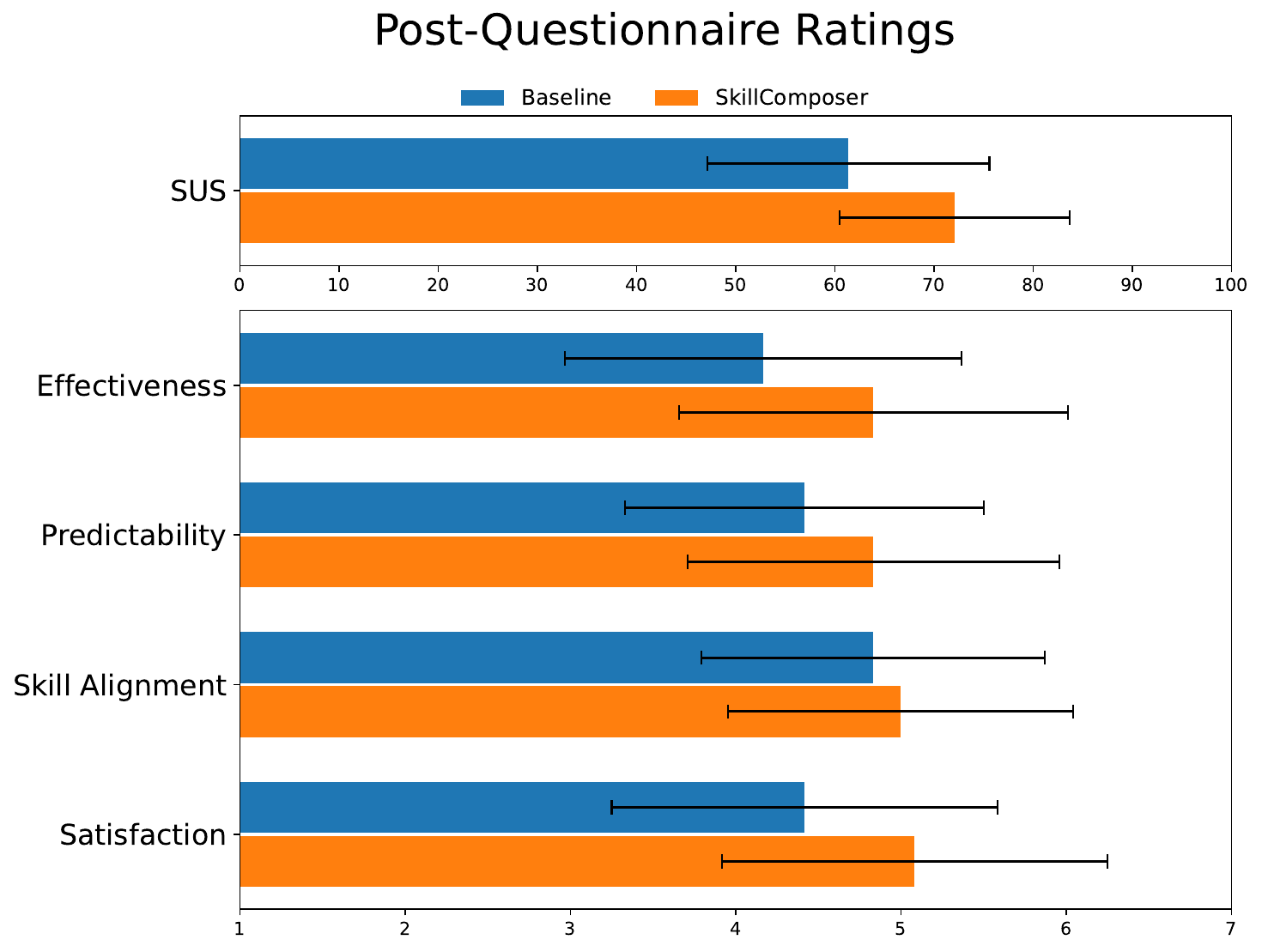}
\caption{Post-questionnaire results from the user study. SUS is measured on a 100-point scale \cite{brooke_sus_1996}, while the other four questions are measured on a 7-point Likert scale, with 7 being the highest. Error bars report $\mu \pm 2\sigma$.}
\label{fig:stats}
\end{figure}

Macro learning primarily affected program complexity rather than success rate.
Systems with macro learning generated shorter programs on average, indicating that learned macros successfully compressed repeated low-level action sequences into higher-level function calls.
Macro learning alone did not substantially improve success over the baseline though, suggesting that learned macros are most effective when paired with the generate-test loop.
Overall, the generate-test system achieved the highest success rate,while the full SkillComposer system had comparable performance
with more compact programs and visible macro reuse.

We also analyzed macro usage over time in SkillComposer.
Figure \ref{fig:macro} shows the average proportion of macros in generated programs across the 20 prompts in each environment, along with the average number of learned macros available at each point.
In both environments, macro availability generally increased as more accepted programs were added to the corpus, and macro usage tended to increase as more learned skills became available.
The learned macros captured recurring action patterns in each environment.
Examples include \verb|move_and_place_object| in the Objects environment, and \verb|place_object_on_plate| and \verb|hand_object_to_user| in the Feeding environment.

\subsection{User Study}

Table \ref{tab:user_results} summarizes the task-level results from the user study.
SkillComposer achieved equal or higher task completion rates than the baseline system on the Rearrange Objects and Meal Preparation tasks while requiring fewer prompts on average across all three tasks.
SkillComposer also reduced average task completion time for the Rearrange Objects and Meal Preparation tasks.
The exception was the Feed Care Recipient task, where the baseline achieved a slightly higher success rate and lower completion time.
This task often required small corrective movements near the end of execution, making rapid interaction more beneficial than higher-level task planning.

The largest difference between the systems was observed in the Rearrange Objects task, where the baseline achieved substantially lower task success.
Participants tended to give high-level instructions involving multiple objects (e.g. P10 prompt \emph{``I want you to swap the positions of the knife and the spoon.''}), whereas the baseline often required users to decompose these requests into basic prompts focusing on one object or action at a time.
As participants gained experience with the baseline system, they gradually adapted their prompting strategies to step-by-step commands, reducing this performance gap in the next two tasks.

Figure \ref{fig:stats} summarizes participants' subjective ratings.
Overall, participants reported a more favorable perception of SkillComposer across all subjective measures.
The largest effect was observed for SUS, where SkillComposer scored 10.7 points higher on average, indicating a moderate effect (Cohen's $d = 0.46$).
SkillComposer achieved a mean SUS score of 72.1 ($\sigma = 20.1$), categorized as good usability \cite{bangor_empirical_2008}.
The baseline only achieved a mean SUS score of 61.4 ($\sigma = 24.6$), categorized as OK usability \cite{bangor_empirical_2008}.
However, paired-samples $t$-tests found no statistically significant difference between systems in any subjective metrics (all $p > 0.05$).

The qualitative responses provide additional insight into these trends.
11 of the 12 participants preferred SkillComposer over the baseline, often noting its stronger support for high-level, multi-step instructions and lower effort.
P7 commented that \emph{``In [SkillComposer], I could use a single prompt. Whereas in [the baseline], I have to break down one task into six or seven prompts to get it to work.''} 
Similarly, P8 said that \emph{``It feels like [SkillComposer] understands me. But the [baseline] feels more like a robot. I have to kind of program a little bit.''} 
Several participants also described adapting their prompting strategies over time.
P11 explained that they initially used \emph{``normal human language''} with SkillComposer, but learned that the baseline required \emph{``very specific prompts.''}
P5 echoed this sentiment, noting that switching from repeated commands to a single high-level prompt \emph{``made my life easier.''}
These observations are consistent with the baseline's higher prompt count and its lower initial success rate on the Rearrange Objects task, where participants were still learning how to prompt the system effectively.

Participants also suggested future improvements and applications.
Several wanted improved accuracy and response time.
Others were interested in applying SkillComposer to additional scenarios, such as household humanoid robots that retrieve objects or bimanual tasks like pouring wine into a cup as P9 suggested.
Participants also suggested interface improvements, including a more visually attractive UI, adding voice interactions, a higher-frame-rate simulation preview, and clearer status notifications.
Together, these comments suggest that participants saw SkillComposer as a promising interface with potential for broader use.

\section{Conclusion}

This paper presents SkillComposer, an interactive natural language robot programming system that combines evaluator-guided code generation with online skill learning. 
By validating generated programs before execution and learning reusable macros from accepted programs, SkillComposer enables the available robot skill set to evolve over time rather than remaining fixed. 
Ablation experiments showed that the generate-test loop substantially improved reliability, while macro learning reduced program length and enabled the reuse of learned skills across tasks. 
A user study further demonstrated that participants generally preferred SkillComposer over a baseline coding LLM, reporting that it better supported high-level, multi-step instructions and required fewer prompts to complete tasks.
As the number of learned skills grows, the coder LLM's ability to utilize these skills may degrade. 
Future work could address this by integrating retrieval-augmented generation to retrieve the most relevant learned skills from longer interaction histories, enabling long-term use with larger skill libraries.
We believe that systems capable of continuously learning from user interaction can make robot programming more accessible by allowing users to describe tasks naturally while the system learns reusable skills tailored to their needs.

\bibliographystyle{ieeetr}
\bibliography{bib}

\end{document}